\documentclass{article}

\usepackage{microtype}
\usepackage{graphicx}
\usepackage{subcaption}
\usepackage{booktabs} 

\usepackage{hyperref}

\newcommand{\vectors}[1]{\lowercase{\mathbf{#1}}}

\usepackage{subcaption}

\usepackage{algorithm}
\usepackage{amsmath}
\usepackage{amsfonts}
\usepackage{enumitem}

\newtheorem{definition}{definition}

\usepackage[accepted]{icml2018}
\icmltitlerunning{Hop-Hop Relation-aware Graph Neural Networks}

\begin{document}

\twocolumn[

\icmltitle{Hop-Hop Relation-aware Graph Neural Networks}

\icmlsetsymbol{equal}{*}

\begin{icmlauthorlist}
\icmlauthor{Li Zhang}{Sheffield}
\icmlauthor{Yan Ge}{Sheffield}
\icmlauthor{Haiping Lu}{Sheffield}
\end{icmlauthorlist}

\icmlaffiliation{Sheffield}{Department of Computer Science, University of Sheffield, Sheffield, UK}

\icmlcorrespondingauthor{Li Zhang}{lzhang72@sheffield.ac.uk}
\icmlcorrespondingauthor{Yan Ge}{yge5@sheffield.ac.uk }
\icmlcorrespondingauthor{Haiping Lu}{h.lu@sheffield.ac.uk}
\icmlkeywords{Machine Learning, ICML}

\vskip 0.3in
]




\begin{abstract}
Graph Neural Networks (GNNs) are widely used in graph representation learning. However, most GNN methods are designed for either homogeneous or heterogeneous graphs. In this paper, we propose a new model, Hop-Hop Relation-aware Graph Neural Network (HHR-GNN), to unify representation learning for these two types of graphs. HHR-GNN learns a personalized receptive field for each node by leveraging knowledge graph embedding to learn relation scores between the central node's representations at different hops. In neighborhood aggregation, our model simultaneously allows for hop-aware projection and aggregation. This mechanism enables the central node to learn a hop-wise neighborhood mixing that can be applied to both homogeneous and heterogeneous graphs. Experimental results on five benchmarks show the competitive performance of our model compared to state-of-the-art GNNs, e.g., up to 13K faster in terms of time cost per training epoch on large heterogeneous graphs.

\end{abstract}

\section{Introduction}
\label{set:intro}

Graphs, such as social networks, knowledge graphs and citation networks, are ubiquitous data structures capturing interactions between individual nodes. They can be broadly divided into homogeneous (one type of nodes and edges) and heterogeneous (multiple types of nodes and edges) graphs \cite{yang2020heterogeneous}. Graph Neural Networks (GNNs) have been widely used in graph representation learning and achieved impressive performance in various applications, such as academic citation networks \cite{kipf2016semi,velickovic2017graph}, social networks \cite{hamilton2017inductive}, knowledge graphs \cite{schlichtkrull2018modeling} and recommender systems \cite{vdberg2017graph,ying2018graph}.

A general GNN framework mainly consists of two key steps: neighborhood aggregation and feature transformation. In neighborhood aggregation, a given node first aggregates (such as sum, mean and pooling) its neighbors, followed by a linear mapping or a multi-layer perceptrons (MLPs) in the feature transformation step to get the node's new representation \cite{hornik1989multilayer}. 

Based on the basic GNN framework, improved GNNs for homogeneous and heterogeneous graphs have been proposed for better network representation learning. The majority of current GNNs assume graphs as HOmogeneous graphs (GNNs-HO), and they usually aggregate central node's one-hop neighbors \cite{kipf2016semi,velickovic2017graph}, neighbors sampled from fixed-length random walks \cite{hamilton2017inductive} or multi-scale neighbors \cite{samimixhop,Liao2018GraphPN,Atwooddnn} in the neighborhood aggregation step. 
In recent years, there are some attempts to apply GNNs to HEterogeneous graphs (GNNs-HE) \cite{schlichtkrull2018modeling,wang2019heterogeneous}, in which different types of nodes are connected under unique relations \cite{yang2020heterogeneous}. Most existing methods aggregate the neighbors from  manually designed or automatically learned meta-paths \cite{wang2019heterogeneous, graphtransformer}.



Two key differences or issues in neighborhood aggregation step of GNNs-HO and GNNs-HE are:
\begin{itemize}
    
    \item \textit{How to define the receptive field}. In GNNs-HO, the receptive field is determined by the powers of adjacency matrix, and it is a meta-path in GNNs-HE. However, the receptive fields in GNNs-HO and GNNs-HE are fixed. In GNNs-HO, some nodes with sparse connection may need to aggregate more (higher-order) neighbors, while one-hop neighborhood information may be enough for some nodes with dense connectivity. The manually defined or learned meta-path is a general receptive field, which may not be suitable for each node in a graph.

    \item \textit{How to aggregate the neighbors}. GNNs-HO treat all neighbors in the receptive field equally in the aggregation. However, different hops of neighbors show different importance in central node's representation learning, e.g., the directly linked (one-hop) neighbors have a closer relationship with the central node than undirectly linked (higher-order) neighbors. Thus, they should be treated differently. GNNs-HE only aggregate the two end nodes in the meta-path and discard all intermediate nodes along the meta-path, which results in information loss. 
\end{itemize}

Can we unify GNNs-HO and GNNs-HE in a general framework while satisfying the following \textbf{D}esirables?


\begin{itemize}

   \item \textbf{D1}: \textit{A personalized receptive field}. The connectivity for each node varies greatly in a graph \cite{velickovic2017graph,murphy2018janossy}. Fixing the aggregation receptive field size for each node (a fixed hops or a general meta-path) tends to result in suboptimal performance. 
 
    \item \textbf{D2 \& D3}: \textit{Hop-aware projection and Hop-aware aggregation}. Different hops or types of neighbors have different traits and their embeddings should be mapped in different feature spaces \cite{schlichtkrull2018modeling}. Besides, these neighbors have different relationships with the central node \cite{samimixhop} so they should be treated differently during aggregation.

    \item \textbf{D4 \& D5:} \textit{Efficiency and interpretability}. For practical applications where efficiency and interpretability are needed \cite{gnnexplainer}, the model should be able to explain which hops or types of neighbors play an important role for the central node's representation learning. Moreover, the model should deal with the complex neighborhood in an efficient way, especially for large size and heterogeneous graphs \cite{dong2020HeterNRL, wang2019heterogeneous}.
\end{itemize}

To this end, we propose a universal GNN approach addressing the above desirables for both homogeneous and heterogeneous graphs by leveraging a key concept from Knowledge Graph Embedding (KGE), relation-score, which captures the interaction between \textit{head} and \textit{tail} entities in a knowledge graph (KG) \cite{wang2017knowledge}. Our model mainly contains three modules: GNN, KGE and concatenation. Motivated by the observation that different types of neighbors along a meta-path in a heterogeneous graph can be treated as different hops of neighbors, we utilize different weight matrices for different hops of neighbors to learn central node’s representations at different hops (\textbf{D2}). In the KGE module, we feed the low-dimensional embeddings from the GNN module to a neural tensor network (NTN) \cite{socher2013reasoning} to simultaneously learn the relation-scores\footnote{For a given node, neighbors in the same hop have the same relation-score.} (e.g., $\alpha_{01}, \alpha_{02}, ..., \alpha_{0p}$) between a central node (e.g., node 0) and its embeddings at different hops (e.g., $1,2,...,p$-hop). This will enable the central node to softly aggregate neighborhood information at different hops (\textbf{D1}) and this can be done efficiently (\textbf{D4}). Finally,
instead of directly concatenating these embeddings, we concatenate a central node's embedding and its embeddings at different hops weighted by their corresponding relation-scores (\textbf{D3}) to get the central node's new embedding. This will enable our model to mix latent information from neighbors at various distances and types. Analysing the learned relation-scores $\alpha_{01}, \alpha_{02}, ..., \alpha_{0p}$ benefits interpretability (\textbf{D5}). 
 
 

In summary, the key contributions of this paper are:
\begin{enumerate}
    \item We propose Hop-Hop Relation-aware Graph Neural Network (HHR-GNN), a new class of GNNs, to unify GNN-based homogeneous and heterogeneous graph representation learning.
    
    \item We comprehensively evaluate the superiority of our model on both homogeneous and heterogeneous graphs for the popular node classification task.
    
    \item We demonstrate the proposed model to be efficient (up to 13K faster in terms of time cost) while providing interpretable explanation for the prediction.
\end{enumerate}

\section{Preliminaries}

In this section, we first give formal notations and definitions of some important terminologies.

A graph with $N$ nodes can be represented as $\mathcal{G}$= $(\mathcal{V},\mathcal{E}, \mathbf{X})$, where node $v_i \in \mathcal{V}$, edges $(v_i, v_j)\in\mathcal{E}$ ($i,j=1,...,N$), and a feature matrix $\mathbf{X} \in \mathbb{R}^{N \times D}$ containing $N$ $D$-dimensional feature vectors. A hidden representation of node $v_i$ learned by the $k$-th layer of a GNN model is denoted by $\mathbf{H}^{(k)}_{i}$ and we initialize $\mathbf{H}^{(0)}= \mathbf{X}$.


\begin{definition}
\label{de:heterogeneous} 
Heterogeneous graph \cite{yang2020heterogeneous}. Heterogeneous graph is a graph with multiple types
of nodes and links, each node is associated with a node type, and each link is associated with a link type. It is worth noting that
the type of a link $ \mathcal{E}_{ij}$ automatically defines the types of nodes $v_{i}$ and $v_{j}$ on its two ends.
\end{definition}

In heterogeneous graph, two nodes can be connected via different semantic paths, which are called meta-paths.
\begin{definition}
\label{de:meta-path}
Meta-path \cite{sun2012mining}. A is a path defined on the network schema in a form of  $v_{1}\stackrel{r_{1}}{\longrightarrow} v_{2} \stackrel{r_{2}}{\longrightarrow} ... \stackrel{r_{p}}{\longrightarrow} v_{p+1}$, where $v$ and $r$ are node types and link types, respectively.
\end{definition}

\section{Proposed Approach}
\label{sec:approach}







We start from a theoretical study of the two types of graphs and their neighborhood aggregation to reveal the relationship between neighborhood aggregations in GNNs-HO and GNNs-HE.
This enables us to derive a general framework that is suitable for both homogeneous and heterogeneous graphs representation learning and satisfies all the desirables mentioned above.

\subsection{Homogeneous and Heterogeneous Graphs}
This section first introduces how to represent the two types of graphs and the relationship between \textit{hops} in homogeneous graph and \textit{types} in heterogeneous graph. 

Heterogeneous graph is associated with a node type
mapping function $f_{v}$: $\mathcal{V}$ $\to $ $\mathcal{T}_{v}$  and a link type mapping function
$f_{e}$: $\mathcal{E}$ $\to $ $\mathcal{T}_{e}$, where $\left|\mathcal{T}_{v}\right|$ + $\left| \mathcal{T}_{e}\right|$ $>$ 1. If both $\left|\mathcal{T}_{v}\right|$ =1 and $\left|\mathcal{T}_{e}\right|$ =1, it is a homogeneous graph with the same type of nodes and edges. The heterogeneous graph can be represented by a set of adjacency matrices $\left\{ \mathbf{A}_{r} \right\}_{r=1}^{R}$ ($R$=$\left| \mathcal{T}_{e}\right|$), and $\mathbf{A}_{r} \in \mathbb{R}^{N\times N}$ is an adjacency matrix where $\mathbf{A}_{r}[i,j]$ is non-zero when there is a $r$-th type edge from $v_{j}$ to $v_{i}$. In homogeneous graph, the adjacency matrix is simplified to $\mathbf{A}\in\mathbb{R}^{N\times N}$ ($R$=1). 

Definition \ref{de:meta-path} defines a composite relation $R = r_{1} \circ r_{2} ... \circ r_{p} $ between node $v_{1}$ and $v_{p+1}$. Given the composite relation $R$, the adjacency matrix of the meta-path can be obtained by multiplications of adjacency matrices as:
\begin{equation}
    \mathbf{A}_{R} = \mathbf{A}_{r_{1}} \mathbf{A}_{r_{2}}...\mathbf{A}_{r_{p}}.
\label{eq:metapath}
\end{equation}
For example, the meta-path Author-Paper-Conference ($APC$) in citation graphs, which can be represented as $A \stackrel{AP}{\longrightarrow}P\stackrel{PC}{\longrightarrow} C$, generates an adjacency matrix $\mathbf{A}_{APC}$ by the multiplication of $\mathbf{A}_{AP}$ and $\mathbf{A}_{PC}$. 
In homogeneous graph, 

\begin{equation}
    \mathbf{A}_{R} = \mathbf{A}^{p},
\label{eq:ho-metapath}
\end{equation}
because $\mathbf{A}_{r_{1}} =  \mathbf{A}_{r_{2}} ... = \mathbf{A}_{r_{p}}$. 

Equations \ref{eq:metapath} and \ref{eq:ho-metapath} illustrate the relation between \textit{ types} in heterogeneous graph and \textit{hops} in homogeneous graph. Therefore, we can view a meta-path as high-order proximity between two nodes. The adjacency matrix $\mathbf{A}_{APC}$ can be viewed as the two-hop connectivity matrix for A type node.

\subsection{Theoretical Studies of Neighborhood Aggregation}
After revealing the relationship between homogeneous and heterogeneous graphs, this section mainly studies the neighborhood aggregation in GNNs-HO and GNNs-HE.

A general GNN layer can be defined as:
\begin{equation}
    \mathbf{H}^{(k)} = \sigma (\mathbf{A}^{p} \mathbf{H}^{(k-1)}\mathbf{W}^{(k)} ),
\label{eq:one-hop-repre}
\end{equation}
where $\sigma$ can be any activation function, $\mathbf{H}^{(k-1)} \in \mathbb{R}^{N\times d_{k-1}}$ and $\mathbf{H}^{(k)} \in \mathbb{R}^{N\times d_{k}}$ is the input and output for layer $k$,  $\mathbf{W}^{(k)} \in \mathbb{R}^{d_{k}\times d_{k-1}}$ is the trainable transformation matrix.

The neighborhood aggregation iteratively updates the representation of a node by aggregating its neighbors' representations. To mathematically formalize the above insight, the aggregation process can be generically written as follows:

\begin{equation}
	\setlength{\abovedisplayskip}{2pt}
	\setlength{\belowdisplayskip}{2pt}
	\vectors{s}^{(k)}_{i} = f_{ag}^{(k)} (\vectors{h}_{j}^{(k-1)}, j \in \mathcal{N}_{i}),
	\label{eq:general aggregate}
\end{equation}
where $f_{ag}^{(k)}$ is the predefined aggregation function in the $k$-th layer of a model, $\mathcal{N}_{i}$ is the defined neighborhood and $\vectors{s}^{(k)}_{i}$ is the aggregation result. One key part is how to define the neighborhood $\mathcal{N}_{i}$. 

In GNNs-HO, $\mathcal{N}_{i}$ is mainly based on the powers of adjacency matrix $\mathbf{A}^{p}$. $\mathcal{N}_{i}$ means one-hop neighbors ($p=1$) or high-order neighbors ($p>1$). Once $p$ is fixed, the receptive fields will also be fixed. 

In GNNs-HE, neighborhood is defined by the manually defined meta-path as Eq. \ref{eq:metapath}. If the meta-path is defined as APA (The APA meta-path associates two co-authors), A type nodes only aggregate A type nodes or its second-order neighbors, so $\mathcal{N}_{i}$ is the second-order neighbors, while the P type (one-hop) neighbors are discarded, resulting in information loss. GTN learns the meta-path by multiplications of softly selected adjacency matrices as following:
\begin{equation}
    \mathbf{A}_{R} = (\alpha_{1}\mathbf{A}_{r_{1}}) (\alpha_{2} \mathbf{A}_{r_{2}}) ... (\alpha_{p}\mathbf{A}_{r_{p}}),
\label{eq:gtn-metapath}
\end{equation}
where $\alpha_{1}, \alpha_{2},..., \alpha_{p}$ are learnable parameters who have to be updated in each training epoch, and $\mathbf{A}_{R}$ in Eq.\ref{eq:gtn-metapath} also needs to be recalculated in each training epoch. This process is very time consuming, especially for large scale graphs. Eq.\ref{eq:gtn-metapath} in GTN  defines a general meta-path (receptive field) for all nodes in a graph, which may not be suitable for each node.

\subsection{Proposed Algorithm}

 \begin{figure}[!t]
	\centering
	\includegraphics[width=0.5\textwidth]{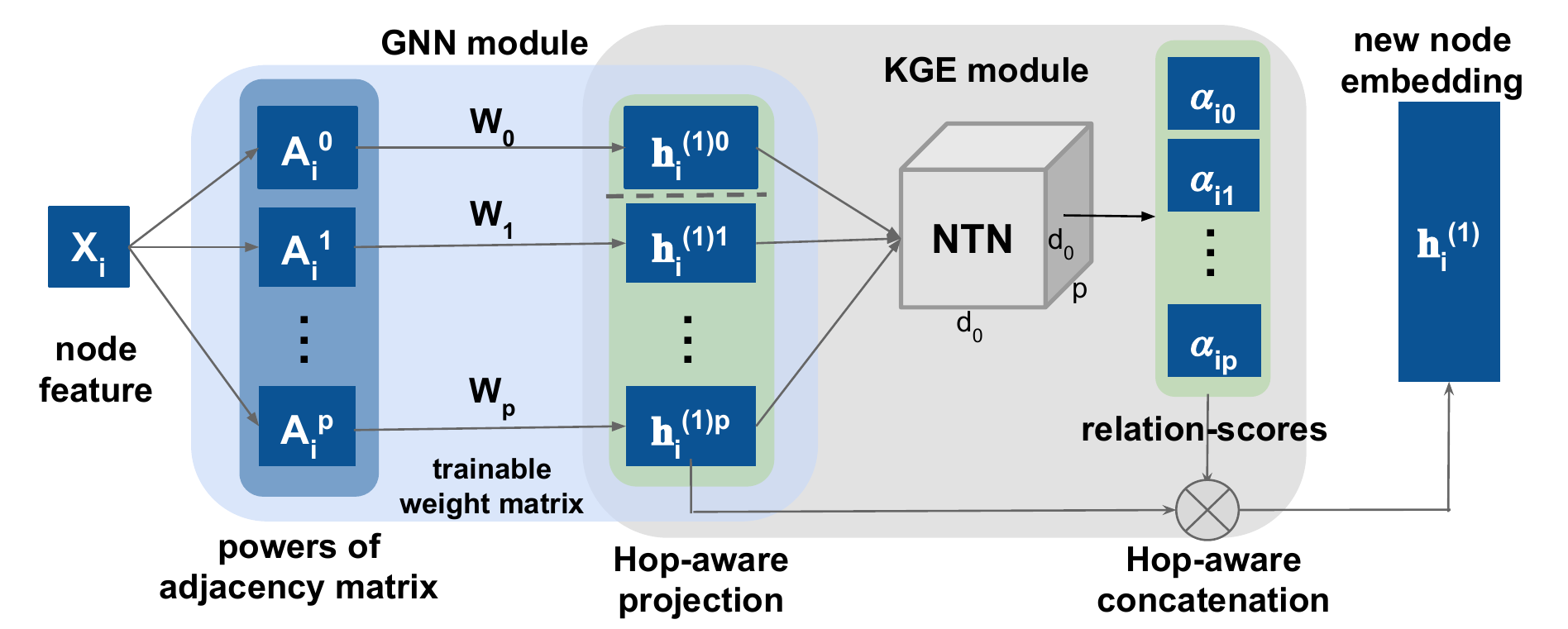}
	\caption{ \small HHR-GNN architecture (the first layer): HHR-GNN first calculates its representations at different hops, e.g., $\vectors{h}^{(1)_{p}}_{i}$ is $p$-hop representation that aggregates neighbors in the $p$-hop. Then the central node's representation $\vectors{h}^{(1)_{0}}_{i}$ and its representations at different hops ($\vectors{h}_{i}^{(1)_{1}}, \vectors{h}_{i}^{(1)_{2}},...,\vectors{h}_{i}^{(1)_{p}}$) will be fed to a NTN model to learn the relation-scores.  Finally, we concatenate each node's embedding with its representations at different hops weighted by their corresponding relation-scores to get the new embedding.}
	\label{fig:algorithms}
\end{figure}


After the theoretical studies, we carefully design our model and propose a new GNN: Hop-Hop Relation-aware Graph Neural Network (HHR-GNN). Our ultimate goal is to design a model that can be suitable for both homogeneous and heterogeneous graphs and satisfy all the desirables. Fig. \ref{fig:algorithms} shows the architecture of our model. GNN module is the hop-aware projection part that maps different hop neighborhood information to different latent space, followed by the NTN in KGE module to learn the relation-scores used for a personalized receptive field. The final step is the hop-aware aggregation, where the central node aggregates different hops embeddings with different attention coefficients based on the learned relation-score in KGE module.

\subsubsection{A Personalized Receptive Field.}
A personalized receptive field (\textbf{D1}) is the foundation for hop-aware projection (\textbf{D2}) and hop-aware aggregation (\textbf{D3}). So, we start from how to design our model for D1.

The purpose of learning a personalized receptive field is that the central node can combine information from different hops neighborhood to assist its representation learning.
Based on the theoretical study and analysis, we need a smarter way to combine neighborhood information from different hops. Instead of learning the combination of adjacency matrices \cite{graphtransformer}, we learn a soft combination of different hops' embeddings as following:

\begin{equation}
	\vectors{h}^{(k)}_{i} = f^{(k)}_{ag} (\alpha_{ir} \vectors{h}_{i}^{(k)_{r}}, r \in (0,1,...,p))
	\label{eq:general aggregate}
\end{equation}
where $\alpha_{ir}$ are learnable parameters and $\vectors{h}_{i}^{(k)_{r}}$ is $v_i$'s $r$-hop representation, $\alpha_{i0} = 1 $.

Eq. \ref{eq:general aggregate} is analogous to the popular receptive module \cite{szegedy2015going} for classic Convolutional Neural Networks (CNN) architectures: it consists of convolutional filters of different sizes determined by the parameter $p$, where $p$ = 0 corresponds to 1 $\times$\ 1 convolutions in the receptive module (amounting to transformations of the features in each node without diffusion across nodes in GNN). The difference is that the convolution filter in CNN convolve all within the $p \times p$ receptive field, and we use $\alpha_{ir}$ to combine different hops information to define a flexible receptive field for each node (\textbf{D1}) within $p$-hop neighborhood. For example, $v_i$ only aggregates its one-hope and three-hop neighbors, when $p=3$ and the learned parameters $\alpha_{i0}, \alpha_{i1}, \alpha_{i3} >0, \alpha_{i2} = 0$.

Two natural follow-up questions: how to learn $\vectors{h}_{i}^{(k)_{r}}$ and $\alpha_{ir}$.



\subsubsection{Hop-aware Projection.}

Considering different hops or types neighbors have different traits and their embeddings should fall in different feature space, we design a \textit{hop-specific} transformation matrix ($\mathbf{W}_{r}$) to project the node features for each hop (or type) nodes (\textbf{D2}). The \textit{$r$-hop representation} can be expressed as:

\begin{equation}
    \vectors{h}_{i}^{(k)_{r}} = \sigma (\mathbf{A}_{i}^{r} \vectors{h}_{i}^{(k-1)}\mathbf{W}^{(k)}_{r}),
\label{eq:hop-type-repre}
\end{equation}
where $\sigma$ can be any activation function, $\mathbf{W}^{(k)}_{r} \in \mathbb{R}^{d_{k}\times d_{k-1}}$ is the trainable transformation matrix, $\vectors{h}_{i}^{(k-1)} \in \mathbb{R}^{d_{k-1}}$ is the hidden representation of the ($k-1$)-th layer and $\mathbf{A}_{i}^{r}$ is the $i$ row or line of  $r$-th power of adjacency matrix $\mathbf{A}$. $\vectors{h}_{i}^{(k)_{r}}$ is the \textit{$r$-hop representation}. 

As for heterogeneous graph, \textit{$r$-hop representation} can also be interpreted as a certain type of neighbors' representation. Because $\mathbf{A}^{r}$ can be seen as the $r$-th step in the meta-path, as defined in Def \ref{de:meta-path}, which also automatically defines the node type (or edge type), as we defined in Def \ref{de:heterogeneous}. 
So, Eq. \ref{eq:hop-type-repre} is a general expression and can be used in both homogeneous and heterogeneous graphs to represent a certain hop or type of representation. Note, if there are different types of nodes in the same hop in heterogeneous graph, we use different transformation matrices to learn the embeddings. For example, there are two types of nodes in the one-hop neighbors, and we will use $\mathbf{W}^{(k)}_{r1}$ and $\mathbf{W}^{(k)}_{r2}$ to mapping the two types of neighbors.

\subsubsection{Relation-score Learning.}

After obtaining representations at different hops, we will illustrate how to calculate $\alpha_{i1}, \alpha_{i2},...,\alpha_{ip}$ that can define a personalized receptive fields as shown in Eq. \ref{eq:general aggregate}. $\alpha_{ir} (r = 1, ..,p)$ should be high if the $r$-hop representation has a close relationship with central node in homogeneous graph, or $r$-type neighbors should be considered more. In other words, $\alpha_{i1}, \alpha_{i2}, ...,\alpha_{ip}$ reflect the central node and its different hops relationship. Therefore, we transfer this problem to how to model the relationship between $\vectors{h}^{(k)_{0}}_{i}$ and $\vectors{h}^{(k)_{r}}_{i}$ (r = 1,2,...,$p$), which inspires us to apply Knowledge Graph Embedding (KGE) method to solve this problem. KGE methods aim to model the relationship between $head$ $entity$ and $tail$ $entity$ in knowledge graph (KG) and assign a score of how likely it is that two entities are in a certain relationship \cite{socher2013reasoning}. In our scenario, we define the relation-score as: 

\begin{definition}
\label{de:relation-score} 
Relation-score. Relation-score is to model the relationship between central node's embedding and a certain hop or type representation, such as $\alpha_{ir}$ = $f_{r}(\mathbf{h}_{i}^{(k)_{0}}, \mathbf{h}_{i}^{(k)_{r}})$ is the relation-score of $v_{i}$ with its $r$-hop representation.
\end{definition}

We want to model the relationship between central node with its $p$ types of representations simultaneously and usually $p>1$. So, we introduce an expressive neural tensor network (NTN) \cite{socher2013reasoning} that can relate two input entities vectors across multiple dimensions and each slice of the tensor is responsible for one type of entity pair. So, after getting the embeddings, we compute the relation-score by the following NTN-based function:

\begin{equation}
	\boldsymbol{\alpha}^{(k)}_{i} = f (\vectors{h}_{i}^{(k)_{0}}  \mathbf{W}_{R}^{[1:p]} \boldsymbol{\chi} ),
	\label{eq:relation score}
\end{equation}
where $f$ is a  nonlinear activation function, $\mathbf{W}_{R}^{[1:p]} \in \mathbb{R} ^{d_{k} \times d_{k} \times p} $ is a tensor and each slice of the tensor is responsible for instantiation of a relation. $p$ types of representations {$\vectors{h}_{i}^{(k)_{1}}, \vectors{h}_{i}^{(k)_{2}},...,\vectors{h}_{i}^{(k)_{p}}$} can be written as $\boldsymbol{\chi} \in \mathbb{R}^{ d_{k} \times p}$.
$\boldsymbol{\alpha}^{(k)}_{i} \in \mathbb{R} ^{p}$ is the learned relation-score vector for each relation type, where each entry is computed by on slice of the tensor $\mathbf{W}_{R}^{[1:p]}$, such as 
\begin{math}
	\alpha^{(k)}_{ir} = f (\vectors{h}_{i}^{(k)_{0}}  \mathbf{W}_{R}^{[r]} \vectors{h}_{i}^{(k)_{r}}).
\end{math}

Eq. \ref{eq:relation score} allows to learn multiple types of relationship simultaneously. It should be note that NTN only needs to calculate the relation-scores of  $\vectors{h}_{i}^{(k)_{0}}$  and $\vectors{h}_{i}^{(k)_{1}}, \vectors{h}_{i}^{(k)_{2}},...,\vectors{h}_{i}^{(k)_{p}}$, these low-dimensional embeddings, not central node with all the nodes (N$_{p}$ is the number of nodes and generally N$_{p}$ $\gg$ $p$) within $p$-hop neighborhood, which is very efficiency (\textbf{D4}). 

\subsubsection{Hop-aware Aggregation.}
After learning the relation-score, we multiply different hop-representations with their corresponding relation-score, then concatenate them with central node's embedding to get the new embedding $\vectors{h}_{i}^{(k+1)}$. This allows for a hop discrimination when central node aggregates different hops' embeddings (\textbf{D3}) and can be written as:

\begin{equation}
     \vectors{h}_{i}^{(k)} = \sigma (\parallel_{r=0}^{p} {\alpha}_{ir}\vectors{h}_{i}^{(k)_{r}}),
\label{eq:concatenate}
\end{equation}
where $\parallel$ denotes column-wise concatenation, $\alpha_{ir}$ is the relation-score for $r$-hop embedding of $v_{i}$. Analysing $\alpha_{ir}$ can show which hops have an important influence to $\vectors{h}_{i}^{(k+1)}$, leading to benefits in interpretability (\textbf{D4}).

After applying components introduced in the previous section, we obtain the final node representation, which can be used in different downstream tasks. For multi-class node classification, $\mathbf{H}^{(k)}$ will be passed to a fully-connected layer with a $softmax$ activation function. The loss function is defined as the cross-entropy error over all labeled examples:
\begin{equation}
\mathcal{L}= -\sum_{l\in\mathcal{V}_l}\sum_{f=1}^F \mathbf{Y}_{lf} \ln \mathbf{H}^{(K)}_{lf}\ , 
\label{eq:loss}
\end{equation}
where $\mathcal{V}_l$ is the set of node indices that have labels and $d_{K}$ is the dimension of output features equaling to the number of classes. $\mathbf{Y}_{lf} \in \mathbb{R}^ {\left| \mathcal{V}_l \right|\times F} $ is a label indicator matrix. With the guide of labeled data, we can optimize the
our model via back propagation and learn the embeddings of
nodes and relation-score. The overall process of HHR-GNN in shown in Algorithm \ref{alg:algorithm}.


\subsubsection{Computational Complexity.}

Two key parts are $p$-hop representation (Eq. \ref{eq:hop-type-repre}) learning and relation-score learning (Eq. \ref{eq:relation score}). The powers of adjacency matrix in Eq. \ref{eq:hop-type-repre} can be easily precomputed, because they do not depend on the learnable model parameters, and this effectively reduces the computational complexity of the overall model. Each type or hop of neighbors share the same weight and the relation-score function is also shared by all nodes in a graph. So, the computation can be parallelized across all nodes. The computational complexity of Eq. \ref{eq:hop-type-repre} and \ref{eq:relation score} is $\mathcal{O}( p \times N \times  d_{k} \times d_{k-1}  + p \times N \times d_{k} \times d_{k} )$. As for memory requirement, it grows linearly in the size of the dataset and we perform mini-batch training to deal with this issue.


\begin{algorithm}[tb]
	\centering
	\caption{The overeall process of HHR-GNN}
	\label{alg:algorithm}
	\begin{algorithmic}
		\STATE {\bfseries Input:} 
		$\mathcal{G}$= $(\mathcal{V},\mathcal{E}, \mathbf{X}$) with $N$ nodes,
		A set of adjacency matrices $\left\{ \mathbf{A}_{r} \right\}_{r=1}^{p}$.
		Feature matrix $\mathbf{X} \in \mathbb{R}^{N \times D}$,
		Labeled nodes $\mathcal{V}_l$,
		Label indicator matrix $\mathbf{Y}_{lf} \in \mathbb{R}^{\left| \mathcal{V}_l \right|\times F}$,
		The number of hops: $p$,
		The number of layers K.
		\STATE {\bfseries Output:} The final embedding $\vectors{h}_{i}^{(K)}$,
		      and $p$-hop relation-scores $\boldsymbol{\alpha}^{(K)}_{i}$.
		\STATE Calculate different powers of adjacency matrix $\mathbf{A}^{1},$$\mathbf{A}^{2},$...,$\mathbf{A}^{p}$.
		
		\FOR {k = 1,2,...,K}
			\FOR{each $v_{i}$ $\in$ $\mathcal{V}_l$}
		 
		      \STATE Calculate $\vectors{h}_{i}^{(k)_0}$, $\vectors{h}_{i}^{(k)_1}$,..., $\vectors{h}_{i}^{(k)_p}$\\
		      \STATE Calculate the relation-scores: $\alpha_{i1}, \alpha_{i2}, ...,\alpha_{ip}$\\
		      \STATE Concatenate the central node and its $p$-hop neighbors embeddings with multiplying the corresponding relation-scores to get $\vectors{h}_{i}^{(k)}$
		    \ENDFOR
		\ENDFOR
		\STATE Calculate
		\begin{math}
		\mathcal{L}= -\sum_{l\in\mathcal{V}_l}\sum_{f=1}^F \mathbf{Y}_{lf} \ln \vectors{h}^{(K)}_{lf}\ .
		\end{math}
		\\
		\STATE Back propagation and update parameters in HHR-GNN.
		
		
	\end{algorithmic}
\end{algorithm}

\section{Related Works}
\label{sec:related works}
Our model draws inspiration from GNNs-HO, GNNs-HE and KGE. we provide a brief overview on all related works.

Most existing GNNs follow a neighborhood aggregation or “message passing” scheme. Graph convolutional network (GCN) is derived from spectral graph convolutions \cite{Shuman2013TheEF,bruna2013spectral,Cho2014LearningPR} and simplifies the K-localized Chebynet \cite{NIPS2016_6081,bronstein2017geometric} by only aggregating one-hop neighbors. Graph Attention Networks (GAT) \cite{velickovic2017graph} learns to assign different edge weights at each layer. However, limiting node's receptive fields to only one-hop neihgbors seems arbitrary \cite{klicpera_diffusion_2019}. Some works generalize GCN to incorporate higher-order neighbors. GraphSAGE \cite{hamilton2017inductive} aggregates neighbors sampled from a fixed-length random work. MixHop and Lanczosnet [18] explore multi-scale information based on powers of adjacency matrix. Graph diffusion convolution (GDC) \cite{klicpera_diffusion_2019} transforms the original adjacency matrix via graph diffusion to indirecly leverage high-order neighborhood information. But all these methods only define a fixed neighborhood, and the information from different hops is not treated differently during the aggregation. Besides, many real-world problems often can not be presented as a single homogeneous graph and most graphs with various types of nodes and edges \cite{graphtransformer,yang2020heterogeneous}, where GNNs-HO can not be applied to directly. 

Recently, some works have attempted to model the heterogeneous graph by using GNNs. RGCN \cite{schlichtkrull2018modeling} and HetGNN \cite{zhang2019heterogeneous} use either distinct linear projection weight or type-specific RNN to encode features for each type of adjacent neighbors, without considering the high-order neighbors. Another type  of  algorithms transform a heterogeneous graph into a homogeneous graph defined by manually meta-paths in Heterogeneous Graph Attention Network (HAN) \cite{wang2019heterogeneous} or learning a soft selection of edge types for generating useful multi meta-paths in Graph Transformer Networks (GTN) \cite{graphtransformer}. However, they discard all intermediate nodes along the meta-path by only considering two end nodes, which results in information loss. Besides, the intrinsic design and implementation in GNNs-HE make them very slow of modeling large-scale heterogeneous graphs.

In this paper, we don't dig deep into Knowledge Graph Embedding (KGE). Generally, KGE is used to learn a scoring function of $head$ and $tail$ entities which evaluates an arbitrary triplet and outputs a scalar to measure the acceptability of this triplet. In our methods, the central node's and its different hops embeddings are corresponding to the $head$ and different $tail$ vectors respectively. We apply the popular and expressive NTN \cite{socher2013reasoning} model to learn the relation-scores between central node and its different hops' embeddings. Exploring different KGE models, such as translational distance models (TransE \cite{bordes2013translating}, TransH \cite{wang2014knowledge}, TransR \cite{lin2015learning}) and semantic matching models (RESCAL \cite{nickel2011three}, DistMult \cite{yang2014embedding}), is an important direction for future work.

\section{Experiments}

\subsection{Datasets}
 We conduct semi-supervised node classification experiments on both homogeneous and heterogeneous graphs. Following \cite{kipf2016semi,graphtransformer}, we split the Train/Validation/Test as for the two types of graphs as shown in Table \ref{tab:datasets}.We report the mean accuracy (for homogeneous graph) and f1 score (for heterogeneous graph) of 15 runs with random weight matrix initialization for all of our experimental results.

\begin{table}
    \caption{Overview of the datasets. (E.T. means edge type)}
    \resizebox{.5\textwidth}{!}{
    \begin{centering}
	\begin{tabular}
		{lrrrccc}
		\toprule
		Dataset & Nodes & Edges & Features & Classes & E.T.\\ 
		\midrule
		Cora     & 2,708  & 5,429  & 1,433 & 7 &  1& \\
		Citeseer & 3,327  & 4,732  & 3,703 & 6 & 1& \\
        \midrule
		DBLP & 18,405 & 67,946 & 334 & 3 &  4& \\
		ACM & 8,994 & 25,922 & 1,902 & 3 &  4 & \\
	    IMDB & 12,772 & 37,288 & 1,256 & 3 & 4 & \\
	   \bottomrule
	\end{tabular}
    \end{centering}
    }
	\label{tab:datasets}
\end{table}

\begin{itemize}
    \item \textbf{Homogeneous graphs.} We consider two commonly used citation network datasets: Cora and Citeseer. Each dataset only contains one type of nodes (documents) and edges (citation). Node features correspond to elements of a sparse bag-of-words representation of a document. Each node has one class label \cite{sen2008collective}.
    \item \textbf{Heterogeneous graphs.} We use two citation network datasets DBLP and ACM, and a movie dataset IMDB.
    DBLP contains three types of nodes (papers (P), authors (A), conferences (C)), four types of edges (PA, AP, PC, CP), and research areas of Authors as labels. ACM contains three types of nodes (papers(P), authors (A), subject (S)), four types of edges (PA, AP, PS, SP), and categories of Papers as labels. Each node in is represented as bag-of-words of keywords in DBLP and ACM. IMDB contains three types of nodes (movies (M), actors (A), and directors (D)), and four types of edges (MA, AM, MD, DM) and labels are genres of Movies. Node features are given as bag-of-words representations of plots.
\end{itemize}

\subsection{Baselines and Experimental Setup}

\textbf{For homogeneous graph}, we choose four mostly related methods, GCN \cite{kipf2016semi}, GAT \cite{velickovic2017graph}, GraphSAGE \cite{hamilton2017inductive} and MixHop \cite{samimixhop}.

In our model, we model the central node with its tow-hop representation for Cora and Citeseer. Throughout experiments, we use the Adam optimizer \cite{kingma2014adam} with learning rate 0.008, and set the regularization parameter to 5 $\times $10$^{-4}$, the dropout is 0.6. We train all models for a maximum of 500 epochs and use early stopping with a patience of 20. We use two layers for Cora, and the dimension for each layer are 32 and 8, with two NTN (32$\times$ 32 $\times$ 2, 8 $\times$8 $\times$2) to learn the relationships between central with its one-hop and two-hop respectively. For Citeseer, we use one layer and set the embedding dimension to 32. We use the same architecture as in the original papers for GCN, GAT, GraphSAGE and MixHop, because these algorithms have many hyperparameters.

\textbf{For heterogeneous graph}, following \cite{graphtransformer}, we compare with conventional network embedding methods: DeepWalk \cite{perozzi2014deepwalk}, metapaht2Vec \cite{dong2017metapath2vec}, and GNN-based methods: GCN \cite{kipf2016semi}, GAT \cite{velickovic2017graph}, HAN \cite{wang2019heterogeneous} and GTN \cite{graphtransformer}. DeepWalk, GCN and GAT is originally designed for homogeneous graphs and we ignore the nodes(edges) types and perform these methods on the whole graph. 

These three graphs have three types of nodes, and two-hop neighborhood already contains all types of nodes in the three graphs. Thus, we model central node with its one-hop and two-hop neighbors relationships: AP, AC(A-P-C), AA(A-P-A); PA, PS, PP (P-A-P, P-S-P)) and MA, MD, MM (M-A-M, M-D-M) for the three datasets DBLP, ACM and IMDB respectively, based on the tasks. It should be emphasized that in heterogeneous graph, we use different weight matrices to learn embeddings for different types of nodes in the same hop. For example, P has two types of neighbors in one-hop neighborhood: A and S in ACM dataset and we use two weight matrices to learn the two types of node's embeddings. We use the following sets of hyperparameters DBLP, ACM and IMDB: 0.5 (dropout rate), 5 $\times $10$^{-4}$ (weight decay), two HHR layer with the hidden dimension 32. The learning rate are 0.004 (DBLP, ACM) and 0.006 (IMDB).

\subsection{Experimental Results}

\subsubsection{Node classification results.}

\begin{table}[!t]
\centering
\caption{Node classification for homogeneous graph.} 
\label{tab:results for node classification}
\begin{tabular} 
	{lcc}
	\hline
	Methods & Cora & Citeseer \\
	\hline
	GCN       & 81.5 $\pm$ 0.42  & 70.3 $\pm$ 0.46  \\
	GAT       & 83.0 $\pm$ 0.70 & 72.5 $\pm$ 0.67\\ 
	GraphSAGE & 82.2 $\pm$ 2.70  & 71.4 $\pm$ 1.70\\
	MixHop    & 81.9$\pm$ 0.62 & 71.4 $\pm$ 0.81 \\ 
	\hline
   Ours & \textbf{83.84 $\pm$ 0.77} &  \textbf{72.74 $\pm$ 0.60}  \\
\hline
\end{tabular}
\label{tab:node classification homogeneous}
\end{table} 

\begin{table}[!t]
\centering
\caption{Node classification for heterogeneous graph.}

\label{tab:results for node classification}
\small
\begin{tabular} 
 	{lccc}
	\hline
	Methods &  DBLP & ACM & IMDB\\
	\hline
	DeepWalk  & 63.18 & 67.42 & 32.08 \\ 
	metapath2vec & 85.53 & 87.61 & 35.21 \\ 
	GCN & 87.30 & 91.60 & 56.89 \\
	GAT & 93.71 & 92.33 & 58.14 \\ 
	HAN   & 92.83 & 90.96 & 52.33 \\ 
	GTN   & 94.18 & \textbf{92.68} & 60.92 \\ 
	\hline
   Ours  & \textbf{94.65 $\pm$ 0.26} & 91.83$\pm$ 0.54 & \textbf{61.67 $\pm$ 0.62} \\
\hline
\end{tabular}
\label{tab:node classification result heterogeneous}
\end{table} 

Based on results in Table \ref{tab:node classification homogeneous} and Table \ref{tab:node classification result heterogeneous}, we conclude that our method is very competitive on both homogeneous and heterogeneous graphs. 

\textbf{For homogeneous graph}, our method leverages an enlarged (two-hop) neighborhood, meanwhile each node has their personalized receptive field, which can provide more and useful information for central node's representation learning. This is especially beneficial for sparse graphs. \footnote{The average node degree for Cora and Citeseer are 4.9 and 3.7 respectively.} 

\textbf{For heterogeneous graph}, the results for other methods come from \cite{graphtransformer}. Our method is about 1.5\% better than GTN  and 9.3\% better than HAN on IMDB dataset. We reason this improvement is caused by that our model can provide a personalized context for each node, and utilize different types of neighbors to learn the new representation for central node. HAN utilizes the manually designed meta-path to generate the homogeneous graph and other types nodes information has been lost, which may damage the central node's representation learning and make the performance unstable. The key idea of GTN is to learn a general meta-path, while this path is not suitable for every node in the graph. We encourage all types of nodes to appear within a fixed-hop neighborhood, meanwhile treat different types of nodes differently in the aggregation. Compared with HAN and GTN, our method can provide a customized receptive field for each node and absorb useful information from other types of nodes.

\subsubsection{Efficiency.}

\begin{figure}[!h]
\centering
  \begin{subfigure}[b]{0.8\linewidth}
	\centering
		\includegraphics[width=0.9\textwidth]{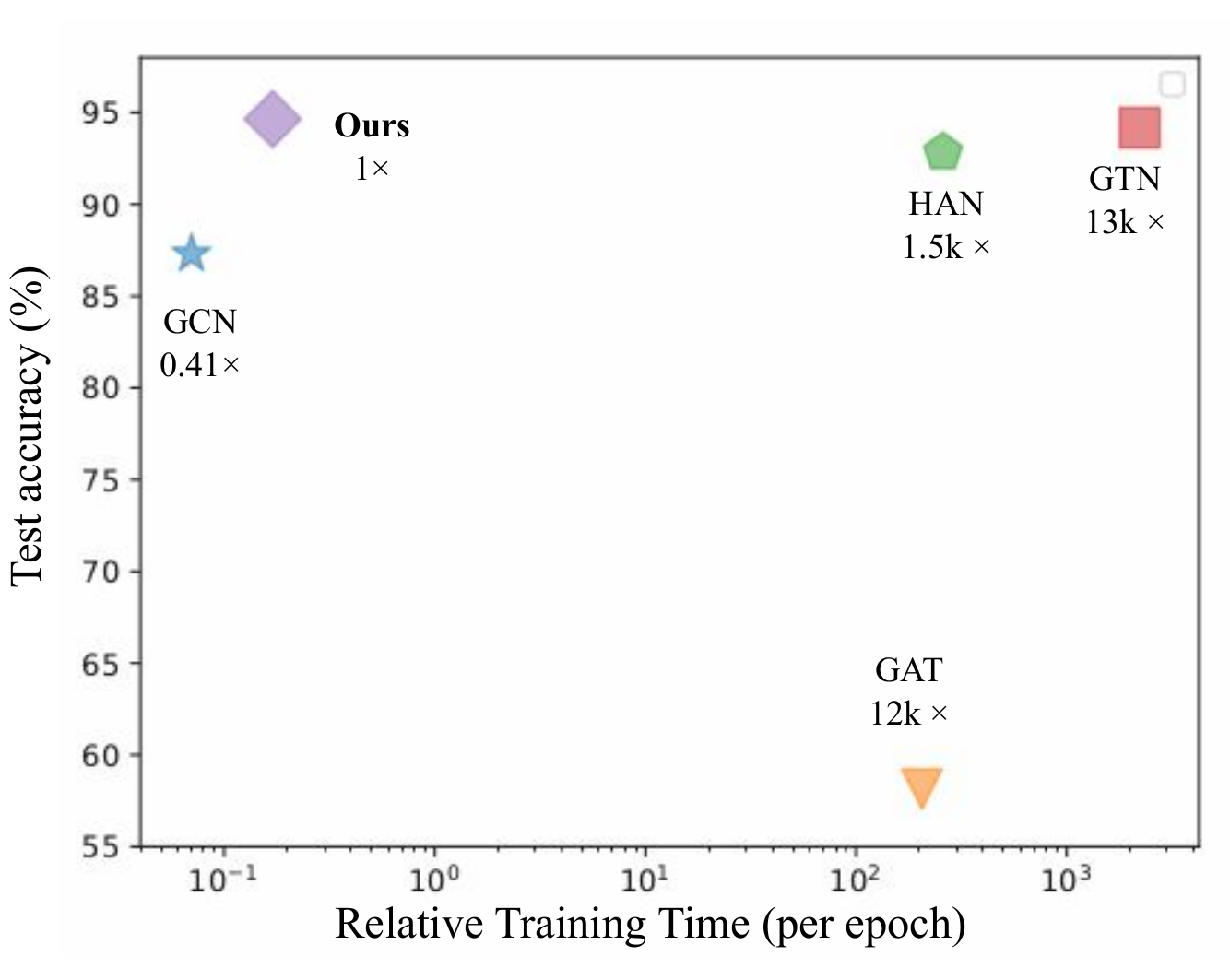}
		\caption{DBLP}
		\label{fig:dblp acc and effeciency}
	\end{subfigure}
	
	\begin{subfigure}[b]{0.8\linewidth}
	\centering
		\includegraphics[width=0.9\textwidth]{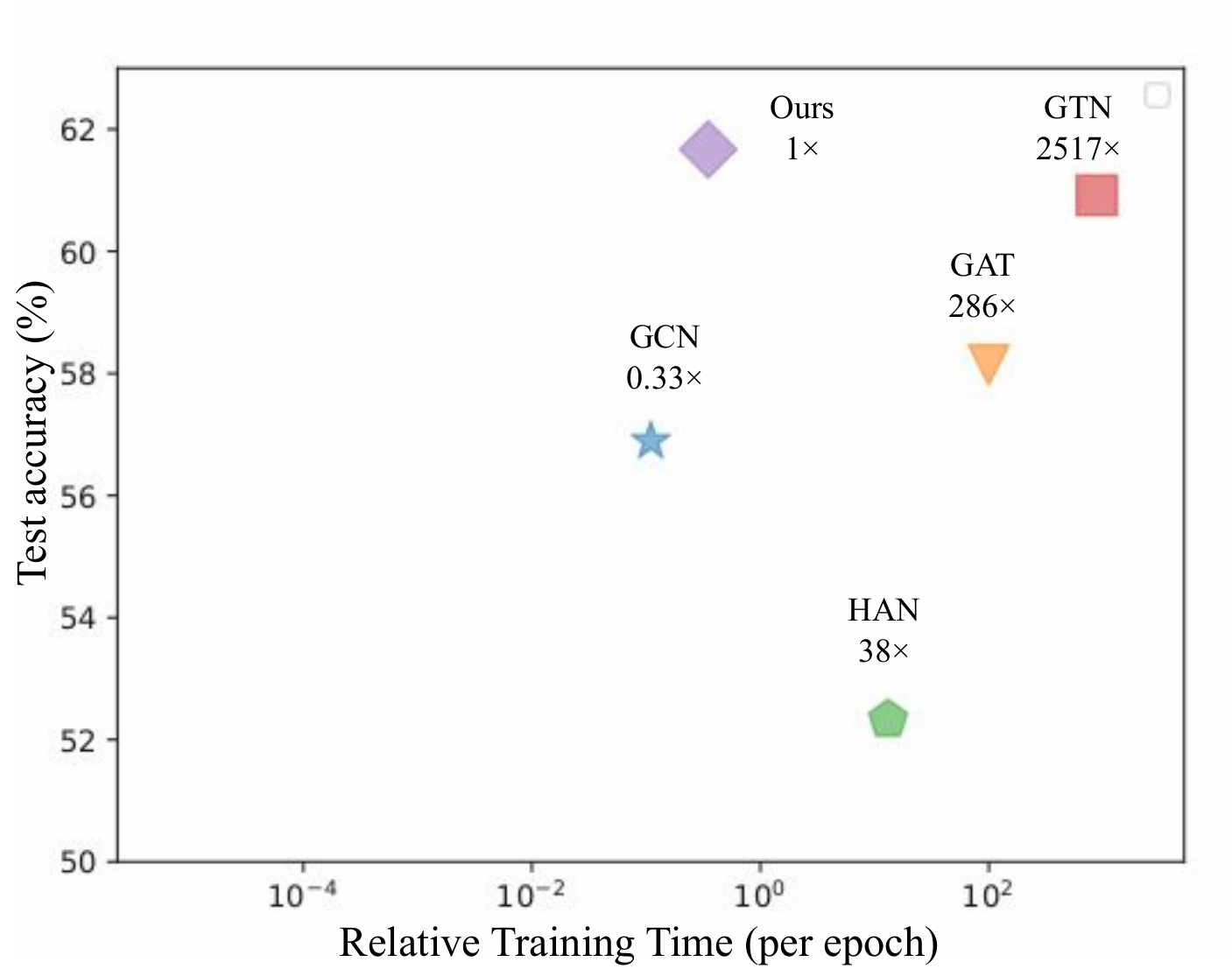}
		\caption{IMDB}
		\label{fig:imdb acc and effeciency}
	\end{subfigure}
	\caption { \small Performance over training time on DBLP and IMDB.}
\end{figure}

In Fig. \ref{fig:dblp acc and effeciency} and Fig. \ref{fig:imdb acc and effeciency}, we plot the performance of the state-of-the-arts GNNs-HE over their training time relative to that of our model on IMDB, and DBLP datasets. The figures shows our model gets competitive performance in both accuracy and efficiently.

Fig. \ref{fig:dblp acc and effeciency} and Fig. \ref{fig:imdb acc and effeciency} show that GCN is the most efficient, but can not ensure the accuracy. Simply aggregating the neighbors is not suitable for heterogeneous graphs that contain more complex neighborhood information than homogeneous graphs. Compared with other methods, GTN has an obvious advantage in accuracy, while does not perform good in efficiency. Because GTN aims to learn 
an optimal meta-path by utilizing a 1 $\times$ 1 convolution to softly select adjacency matrices, which makes the final adjacency matrices are very dense and later computations (multiplication of the selected matrices and GCN operation) are very time consuming, especially for large graphs. While our method learns the meta-path by utilizing the low-dimension hidden representations and a light-weight NTN model, which is very efficient, especially for heterogeneous with a few types of nodes. For heterogeneous graphs with more types of nodes and relations, our model need more mapping matrices and slices of the NTN model, and this will slow the computation, which we will future explore in the future works.

\begin{figure}[!t]
\centering
  \begin{subfigure}[b]{0.8\linewidth}
	\centering
		\includegraphics[width=\textwidth]{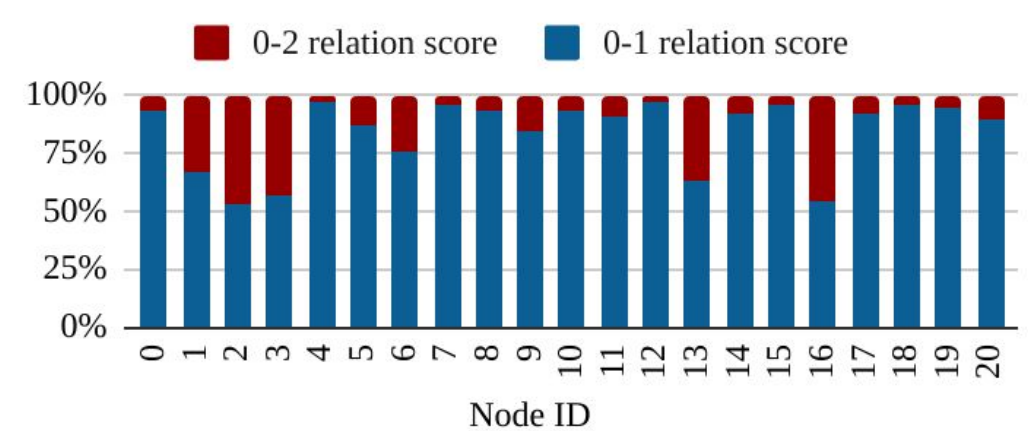}
	     \caption{Cora.}
		\label{fig:cora relation score}
	\end{subfigure}
	\begin{subfigure}[b]{0.8\linewidth}
		\includegraphics[width=\textwidth]{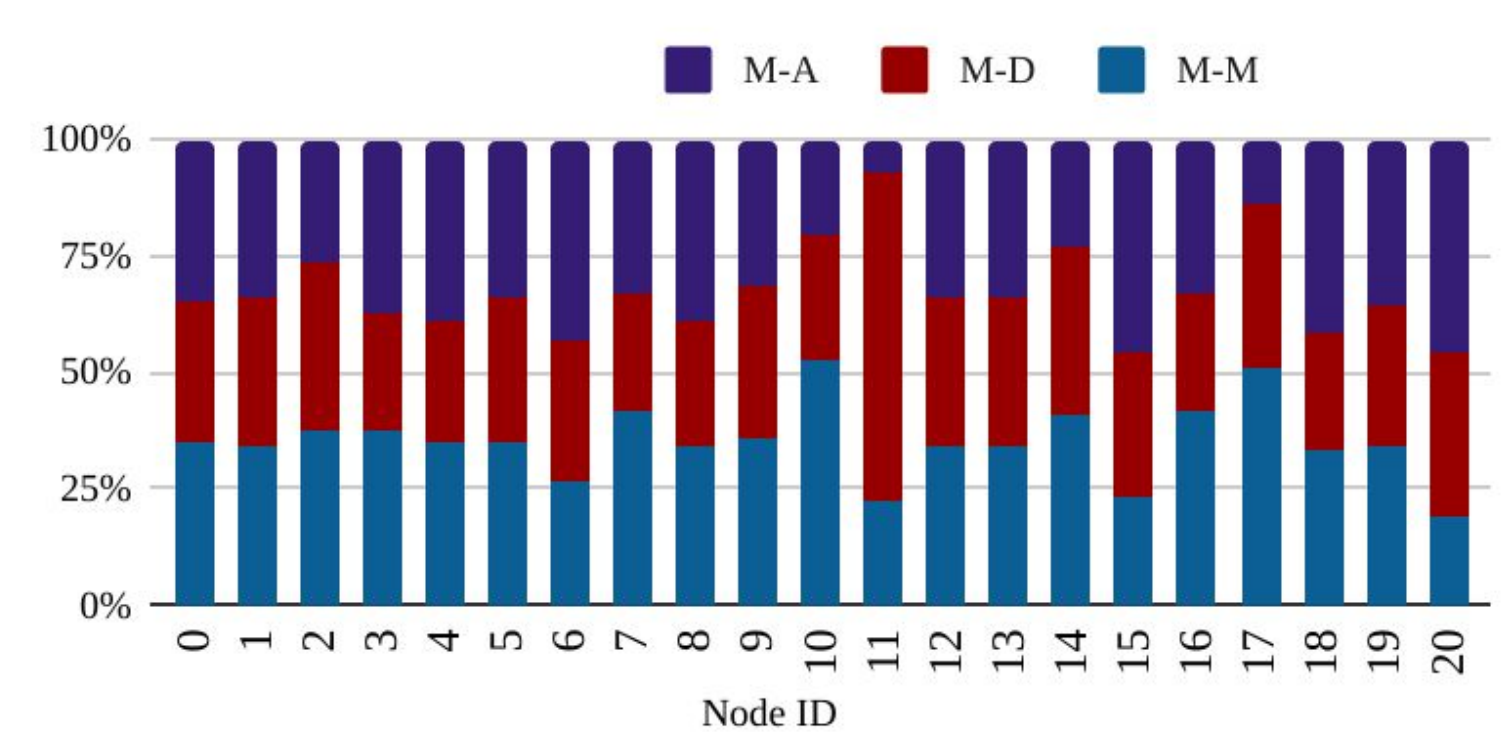}
		\caption{IMDB.}
		\label{fig:imdb relation score}
	\end{subfigure}
	\caption{Cora $0-1$ relation-score and $0-2$ relation-score. IMDB MA, MD and MM relation-score.}
	\vspace{-0.5cm}
\end{figure}

\subsubsection{Interpretability.}

A key part of our model is the hop-aware aggregation by the learned relation-scores.

We first visualize the one-hop (0-1 relation-score ) and two-hop (0-1 relation-score ) of the first 20 nodes on Cora, as shown in Fig. \ref{fig:cora relation score}. For a better comparison, we use the 100\% stacked column chart and more proportions means higher relation-score in each bar. Fig. \ref{fig:cora relation score} shows that 0-1 relation-score is generally higher than 0-2 relation-score, which verifies that directly linked (one-hop) neighbors have more closed relationship than indirectly connected (two-hop) neighbors.

Besides, we also show the learned relation-score of the first 20 nodes on IMDB, who contains four types of edges (MA, AM, MD, DM). We model the relationships of MA, MD (two types of one-hop neighbors) and MM (two-hop neighbors that formed by M-A-M and M-D-M) and visualize the three types of relation-scores in Fig \ref{fig:imdb relation score}. Compared with Fig \ref{fig:cora relation score}, the relation-score is much more complexed, due to the central node is connected by different types of nodes. Fig \ref{fig:imdb relation score} shows that each node has their preference for the different types of nodes, e.g., Node11 has a much closer relationship with Director (the red part takes a big proportion in the bar), while Node10 with Movie.

 \section{Conclusion}
We proposed Hop-Hop Relation-aware Graph Neural Networks (HHR-GNN), a new class of Graph Neural Networks, that can be used for both homogeneous and heterogeneous graphs' representation learning. We introduced the knowledge graph embedding technical to learn the relation-score that can be used to define a personalized receptive fields for each node and applied both hop-aware projection and aggregation to distinguish different hops or types nodes in neighborhood aggregation. We evaluated HHR-GNN against state-of-the-art GNNs on node classification task. Experimental results showed that HHR-GNN is competitive no matter in accuracy and efficiency. Besides, it can identify the useful and personalized context  for each node, which leads to benefits in interpretability and provides insight on the effective hops or types neighbors for prediction.
\nocite{langley00}

\bibliography{mybibliography}
\bibliographystyle{icml2018}

\end{document}